\documentclass[letterpaper, 10 pt, journal, twoside]{IEEEtran}
\IEEEoverridecommandlockouts
\usepackage{cite}
\usepackage{amsmath,amssymb,amsfonts}
\usepackage{graphics}
\usepackage{epsfig}
\usepackage{mathptmx}
\usepackage{times}
\usepackage{wasysym}
\usepackage{upgreek}
\usepackage{placeins}
\usepackage{algorithmic}
\usepackage{graphicx}
\usepackage{textcomp}
\usepackage{xcolor}
\usepackage{siunitx}
\usepackage{booktabs}
\usepackage{multirow}
\usepackage{amsmath, bm}

\def\BibTeX{{\rm B\kern-.05em{\sc i\kern-.025em b}\kern-.08em
    T\kern-.1667em\lower.7ex\hbox{E}\kern-.125emX}}
\usepackage{balance}
\begin{document}

\title{Using R-functions to Control the Shape of Soft Robots}
\author{Declan Mulroy, Esteban Lopez, Matthew Spenko \emph{senior member IEEE}, and Ankit Srivastava%
\thanks{Manuscript received: February, 24, 2022; Revised May, 23, 2022; Accepted June, 20, 2022.}
\thanks{This paper was recommended for publication by Editor Cecilia Laschi upon evaluation of the Associate Editor and Reviewers' comments.
This work is supported by NSF EFRI Grant No. \#1830939 and the Thomas and Josette Morel Graduate Fellowship.
(\textit{Corresponding author is D.\ Mulroy.})
} 
\thanks{All authors are with the Mechanical, Materials and Aerospace Department, Illinois Institute of Technology, Chicago, IL 60616, USA.
        {\tt\footnotesize dmulroy@hawk.iit.edu}}%
\thanks{Digital Object Identifier (DOI): see top of this page.}
}

\markboth{IEEE Robotics and Automation Letters. Preprint Version. Accepted June, 2022}
{Mulroy \MakeLowercase{\textit{et al.}}: Using R-functions to Control the Shape of Soft Robots} 

\maketitle

\begin{abstract}
In this paper, we introduce a new approach for soft robot shape formation and morphing using approximate distance fields. 
The method uses concepts from constructive solid geometry, R-functions, to construct an approximate distance function to the boundary of a domain in $\Re^d$.
The gradients of the R-functions can then be used to generate control algorithms for shape formation tasks for soft robots. 
By construction, R-functions are smooth and convex everywhere, possess precise differential properties, and easily extend from $\Re^2$ to $\Re^3$ if needed. 
Furthermore, R-function theory provides a straightforward method to creating composite distance functions for any desired shape by combining subsets of distance functions.
The process is highly efficient since the shape description is an analytical expression, and in this sense, it is better than competing control algorithms such as those based on potential fields. 
Although the method could also apply to swarm robots, in this paper it is applied to soft robots to demonstrate shape formation and morphing in 2-D (simulation and experimentation) and 3-D (simulation).
\end{abstract}


\IEEEpeerreviewmaketitle
\section{Introduction}\label{sec:introduction}

\begin{figure}
    \centering
    \includegraphics{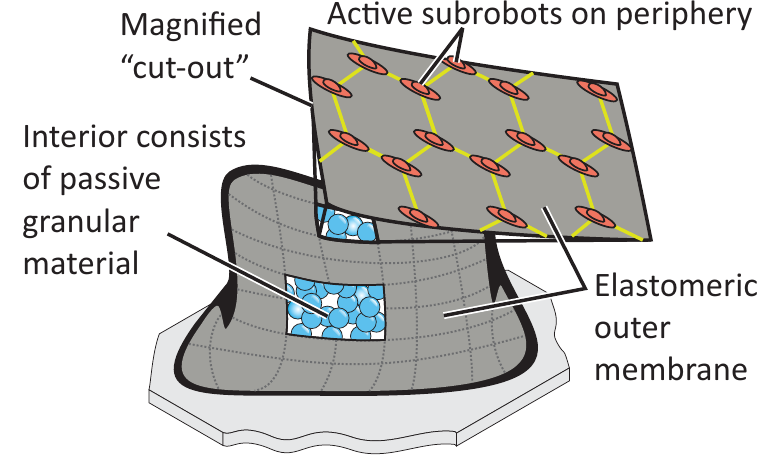}
    \caption{Illustration of three-dimensional boundary constrained soft robot. The robot consists of an elastic membrane enclosing granular material.  On the elastic membrane exists ``active'' sub-robots with the ability to locomote in the environment. Examples of experimental prototypes of boundary constrained swarms can be found in \cite{tanaka2020cable, karimi2020boundary, Karimi2021, agrawal2020scale}. }
    \label{fig:General_Concept_v2}
\end{figure}

\IEEEPARstart{T}{his} paper demonstrates how to incorporate distance functions and image morphing techniques (transfinite interpolation~\cite{Sanchez2015}) to derive control algorithms for soft robots performing both shape formation and shape morphing tasks.
Distance functions have been employed in modeling solids \cite{pottmann2005industrial}, mesh generation \cite{persson2004simple}, topology optimization \cite{osher2003signed}, SLAM and path planning applications~\cite{Fossel2015}, dynamic system planning~\cite{Billard_DS_planning}, and rendering and animation \cite{jeremias2013shadertoy}.
In this work, the distance functions are implemented via R-functions \cite{Shapiro:1991}.
Compared to other techniques, such as radial-basis functions that require the use of tedious numerical algorithms, R-functions provide an analytical solution. 
As a proof of concept, we apply these techniques to a soft robot based on a boundary constrained granular swarm (see Fig.\ \ref{fig:General_Concept_v2}). 

A boundary constrained granular swarm robot is composed of a closed-loop series of active sub-robots, each with the ability to locomote \cite{tanaka2020cable, karimi2020boundary, Karimi2021, agrawal2020scale}.
Each sub-robot is connected to its neighbors with an elastic membrane, and the whole forms a single robot.
In the case of the studies cited above, the membrane encloses a passive granular interior, which provides structure and allows the robot to switch between rigid and soft states via granular jamming phase transitions.

Prior work has demonstrated that robotic swarms (including traditional, non-boundary-constrained systems) can form desired shapes through a potential field-based control algorithm by using the gradients of the potential field to drive the swarm robots to a desired contour, $S$.
This desired curve is embedded into potential fields as the minima of the field.
One example was in \cite{chaimowicz2005controlling}, which created the potential fields using radial basis functions \cite{Ge2004}.
However, creating complex shapes is difficult using radial basis functions; distance functions can be employed as an alternative.
Distance functions have some of the same desired properties as potential fields, the most important of which is that they attain a zero-level set over a prescribed curve or surface, $S$. 
In general, distance functions are created using one of several numerical techniques such as surface interpolation, multiple-object averaging, spatially weighted interpolation, or a distance transform~\cite{payne1992, Borgefors1996}.
For this work, since it is beneficial to retain distance-like properties of the distance functions, we focus on the creation and use of approximate distance functions via R-functions \cite{Shapiro:1991, Shapiro:2007:SAG, Rvachev:1995:RBV}.
The result is a single analytical equation rather then a numerical representation. 

To transform boundary constrained granular swarms from an initial shape, $S_{i}$, to a final shape, $S_{f}$, we will use image morphing techniques. 
Image morphing has previously been applied to unmanned aerial vehicle control algorithms for obstacle avoidance~\cite{Keshmiri2018}. 
However, they have not been applied to shape formation tasks.
To accomplish shape transformation tasks, we used space-time transfinite interpolation as proposed in~\cite{Sanchez2015}. 
This method incorporated R-function theory and transfinite interpolation between two pre-described curves to morph from an initial curve $S_{i}$, to a final curve, $S_{f}$, represented as distance function. 

In summary, this paper describes how to apply approximate distance functions to soft robots in order to form desired shapes in 2D and 3D. 
Additionally, the paper shows the application of space-time transfinite interpolation, originally conceptualized and proposed for image morphing, to accomplish shape morphing objectives.
The techniques are demonstrated on both simulated and experimental boundary constrained soft robots.

The paper is organized as follows.
Section~\ref{sec:distance-functions_R-function} provides background information on distance functions and R-functions.
Section~\ref{sec:Morphing} discusses the morphing algorithm and Section~\ref{sec:Material and Methods} presents the control algorithms, the experimental platform, and the simulation platform.
Section~\ref{sec:Shape-forming} presents the results of a simulated and experimental system forming and morphing between arbitrary shapes.
Finally, Section~\ref{sec:discussion} provides a discussion of the results. 

\section{Distance Functions and R-Functions}\label{sec:distance-functions_R-function}This section provides a brief background on distance functions and R-functions. 
We use well-known distance functions to represent features of the desired curve, $S$. 
R-function theory is used to combine the features into a single analytical distance function that can be used by a soft robot to form a desired shape.

\subsection{Distance Functions}
\label{sec:distance-functions}
A distance function is an implicit representation for curves and surfaces.
To define a distance function, first let $S \subset \Re^d$ denote an object with boundary $\partial S$.
The exact distance function $d(\bm{x})$ gives the shortest distance between any point $\in\Re^d$ to $\partial S$.
Therefore, $d(\bm{x})$ is identically zero on $\partial S$.

Exact distance functions have several drawbacks. 
First, computing the exact distance function to an arbitrary object is computationally expensive.
Second, exact distance functions do not possess continuous derivatives on the medial axis of an object.
Since the objective in this work is to use distance functions in gradient-based control algorithms, their derivatives must be continuous and thus exact distance functions are not appropriate.
Instead, approximate distance functions (formally represented by $\phi(\bm{x})$) that have a closed-form expression and are computationally more efficient are a more viable solution.

For a point $\in \Re^d$ on $\partial S$, it is essential that any approximation to the distance function satisfy $\phi=0$.
$\phi_i \equiv \phi_i(\bm{x})$ is used to denote the approximate distance function of each piece-wise element for $\partial S$ in $\Re^2$ and $\Re^3$.
Features of most shapes are represented using a combination of simple distance functions described by four piece-wise elements: a circle, a line, a sphere, and a plane.
A description of the distance function for each of these is as follows. 

\begin{figure}
    \centering
    \includegraphics[width=0.48\textwidth]{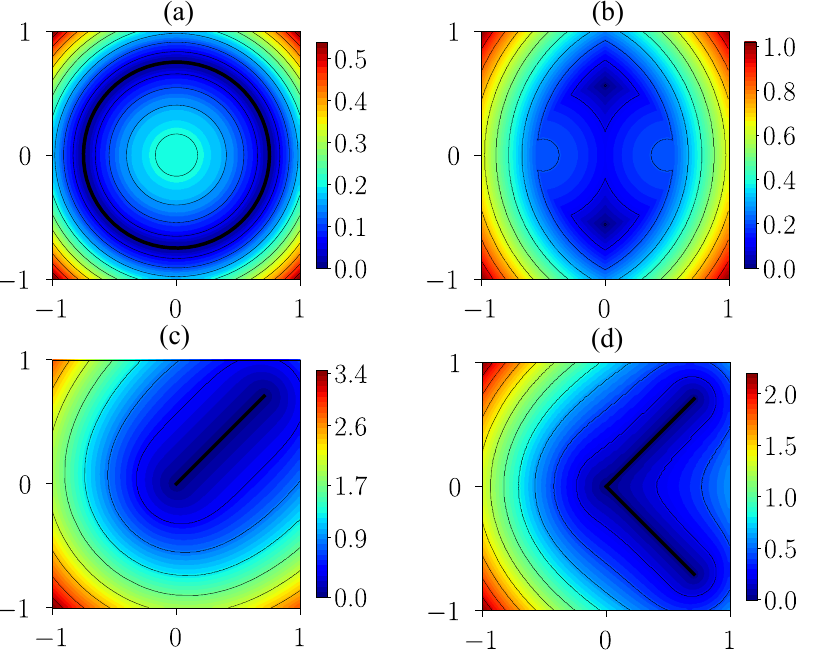}
    \caption{(a): Distance function for a circle of radius 0.75. (b): Union of two circles, each with radius 0.75, centered at $\bm{x}=[0,0.5]$ and $\bm{x}=[0,-0.5]$. (c): Distance function of a line. (d): R-Equivalence of two line segments with $m=2$.}
    \label{fig:dist_r_function}
\end{figure}
\subsubsection{Circle}
The approximate distance function for a circle, $\phi_{c}(\bm{x})$, with a radius R and center $\bm{x}_{c}$ is given as:
\begin{equation}
    \phi_{c}(\bm{x})=\frac{R^{2} - (\bm{x}-\bm{x}_{c})\cdot(\bm{x}-\bm{x}_{c})}{2R}
    \label{eq:circle}
\end{equation}
A visual of a approximate distance function for a circle is shown in Fig.~\ref{fig:dist_r_function}(a).

\subsubsection{Line}
The distance function for a line with end points $\bm{x}_{1}=[x_{1},y_{1}]$, and $\bm{x}_{2}=[x_{2},y_{2}]$ was defined in \cite{Shapiro1999} as:

\begin{equation}
\phi_{l}(\bm{x})=\sqrt{f(\bm{x})^{2}+\left(\frac{\varphi(\bm{x}) -t(\bm{x})}{2}\right )^{2}}
\label{eq:line}
\end{equation}
where $f(\bm{x})$, $t(\bm{x})$ and $\varphi(\bm{x})$ are defined as:

\begin{equation}
f(\bm{x})=\frac{(x-x_{1})(y_{2}-y_{1})-(y-y_{1})(x_{2}-x_{1})}{L}
\label{eq:sdfline}
\end{equation}

\begin{equation}
    t(\bm{x})=\frac{1}{L}\left[ \left(\frac{L}{2}\right )^{2} -||\bm{x}-\bm{x}_{c}||^{2} \right]
    \label{eq:trim}
\end{equation}

\begin{equation}
\varphi(\bm{x})=\sqrt{t(\bm{x})^{2} +f(\bm{x})^{4}}
\end{equation}
where $L=||\bm{x}_{2}-\bm{x}_{1}||_{2}$ and $\bm{x}_{c}=(\bm{x}_{1}+\bm{x}_{2})/2$.
A visual of a approximate distance function for a line is shown in Fig.~\ref{fig:dist_r_function}(c).

\subsubsection{Sphere}
The distance function of a sphere centered at $\bm{x}_{o}$ with a radius $R$ is described as:
\begin{equation}
    \phi_{s}(\bm{x})=(\bm{x}-\bm{x}_{o})\cdot(\bm{x}-\bm{x}_{o}) - R^{2} 
    \label{eq:sphere}
\end{equation}

\subsubsection{Plane}
The distance function for a plane is given as:
\begin{equation}
    \phi_{p}(\bm{x})=(\bm{x}-\bm{x}_{o})\cdot \bm{n}
    \label{eq:plane}
\end{equation}
where $\bm{x}_{o}$ is a known position on the plane and $\bm{n}$ is a vector perpendicular to the plane.
For making curves or surfaces, the approximate distance functions for a circle or sphere are subjected to a trimming function as shown in \cite{Shapiro1999}.


\subsection{R-Functions}
\label{sec:R-function}
R-functions are the tools that combine multiple distance functions to create a single distance function for complex shapes beyond piece-wise elements such as those described in the previous section. 
The R-functions, first proposed by T.\ L.\ Rvachev in 1963~\cite{Shapiro:1991}, construct a composite approximate distance functions, $\phi(\bm{x})$, to any arbitrarily complex boundary, $\partial S$, when approximate distance functions, $\phi_i(\bm{x})$, to the partitions of $\partial S$ are known.

A function, $F(\cdot)$, is an R-function given the following.
Consider a real-valued function $F(\omega_1,\omega_2,\ldots,\omega_q)$, where $\omega_i(\bm{x}): \Re^d \rightarrow \Re$ ($i=1,\ldots,q$) are also real-valued functions.
If the sign of $F(\cdot)$ is solely determined by the signs of its arguments, $\omega_i(\bm{x})$, then $F(\cdot)$ is known as an  R-function~\cite{Shapiro:1991,Shapiro:2007:SAG,Rvachev:1995:RBV}.
R-functions can be used to implicitly describe geometry by utilizing set-theoretic Boolean operations.
R-functions can be combined using techniques similar to a Boolean operation (referred to as R-negation, R-disjunction, and R-conjunction) or through what is known as a R-equivalence operation. 
Each is described below.

\subsection{Disjunction and Conjunction Operations}

Just as Boolean functions are written using the symbols $\neg$ (complement), $\vee$ (union), and $\wedge$ (intersection), every R-function can be written as the composition of the corresponding elementary R-functions: R-negation ($ -\omega$), R-disjunction ($\omega_1 \vee \omega_2$), and R-conjunction ($\omega_1 \wedge \omega_2$).
When defining R-functions for regions in $\Re^d$, a solid can then be composed using the set-theoretic operations of $\neg$, $\vee$, and $\wedge$.
In $\mathbb{U} = \Re^2$, the simplest examples of R-functions are the R-disjunction (union) and the R-conjunction (intersection) functions, given respectively as:
\begin{equation}
R_s(\omega_1,\omega_2) := \frac{ \omega_1 + \omega_2 \pm \sqrt{\omega_1^2 + \omega_2^2 - 2s \omega_1 \omega_2 }}{1+s} 
\label{eq:R_alpha}
\end{equation}
where the $(+)$ and $(-)$ signs define R-disjunction and R-conjunction, respectively, and $s\geq 0$. 
If $\omega_1$ and $\omega_2$ are positive, then so are $\omega_1 \vee \omega_2 $ and $\omega_1 \wedge \omega_2 $.
Fig.~\ref{fig:dist_r_function}(b) provides an example of an R-disjunction operation performed on two circles.

The R-functions defined in Eq.~\ref{eq:R_alpha} are not analytic at points where $\omega_1 = \omega_2 = 0$. 
Smoothness can be obtained by defining the function (where $s=0$ is selected)~\cite{Shapiro:2007:SAG}.
Note that although \emph{R-disjunction} and \emph{R-conjunction} can be used to create approximate distance functions in both $\Re^2$ and $\Re^3$, they are not associative.
A superior method, R-equivalence, is associative and described in the next section~\cite{Millan2015:CBM}.

\subsection{R-Equivalence Operation}\label{subsec:Requiv}
Given two normalized distance functions $\phi_1$ and $\phi_2$ for two curves $S_1$ and $S_2$, a distance field $\phi(\phi_1,\phi_2)$ for the union $S_1 \cup S_2$ must be zero when either $\phi_1 = 0$ or $\phi_2 = 0$ and positive otherwise. 
An R-equivalence solution that preserves normalization up to order $m$ of the distance function at all regular points (nonvertices for polygonal curves) is given by~\cite{Biswas:2004:ADF}:
\begin{equation}
\phi(\phi_1,\phi_2) :=\dfrac{\phi_1 \phi_2}{\sqrt[m]{\phi_1^m +\phi_2^m}}
\label{eq:phi_eq}
\end{equation}
When $\partial S$ (closed curve) is composed of $n$ pieces, then a $\phi$, that is normalized up to order $m$, is described as:
\begin{equation}
\phi(\phi_1,\dots, \phi_n) :=  \dfrac{1}{\sqrt[m]{\frac{1}{\phi_1^m}+\frac{1}{\phi_2^m} + \ldots + \frac{1}{\phi_n^m}}}
\label{eq:phin_eq}
\end{equation}

An example of R-equivalence is shown in Fig.~\ref{fig:dist_r_function}(d) of two lines joined with R-equivalence operation.
Since \emph{R-equivalence} is associative~\cite{Biswas:2004:ADF}, we do not need to consider the order in which the functions are joined to obtain the desired distance function $\phi$.

In summary, we demonstrated two techniques to combine four piece-wise distance functions: point, line, circle, and plane, to create a single analytical expression for a distance function, $\phi$, for an arbitrary shape.
This is superior to other numeric techniques because the R-functions provide an analytical description of a required field that can be computationally efficiently updated. 
The next section explores how a robot might transform from one shape to another.  

\section{Space-Time Transfinite Interpolation}\label{sec:Morphing}
The approximate distance functions provide a gradient field used to drive a boundary-constrained granular swarm soft robot to a particular shape, but in order to transform from an initial shape, $S_{i}$, to a final shape, $S_{f}$, intermediate shapes between the two must be created. This is accomplished using space-time transfinite interpolation~\cite{Sanchez2015}, which is a technique used in image morphing applications.
Space-time transfinite interpolation is given as: 
\begin{equation}
\phi(\bm{x},t)= w_{1}(\bm{x},t)\phi_{i}(\bm{x})+w_{2}(\bm{x},t)\phi_{f}(\bm{x}) 
\label{eq:transfinite_morph}
\end{equation}
where $\phi_{i}(\bm{x})$ and $\phi_{f}(\bm{x})$ correspond to the approximate distance function for the initial and final shapes and $w_{1}(\bm{x},t)$ and $w_{2}(\bm{x},t)$ are weight functions given by:
\begin{equation}
w_{1}(\bm{x},t)=\frac{g_{2}(\bm{x},t)}{g_{1}(\bm{x},t)+g_{2}(\bm{x},t)} 
\label{eq:weight1}
\end{equation}
and
\begin{equation}
w_{2}(\bm{x},t)=\frac{g_{1}(\bm{x},t)}{g_{1}(\bm{x},t)+g_{2}(\bm{x},t)}
\label{eq:weight2}
\end{equation}
where $g_{1}(\bm{x},t)$ and $g_{2}(\bm{x},t)$ are defined as:
\begin{equation}
g_{1}(\bm{x},t)=R_{s}(\phi_{i},-f(t))
\label{eq:g1}
\end{equation}
and
\begin{equation}
g_{2}(\bm{x},t)=R_{s}(\phi_{f},f(t)-1)
\label{eq:g2}
\end{equation}
where $R_{s}(\cdot)$ is the R-conjunction function described in Eq.~\ref{eq:R_alpha}, with $s=0$ and the function $f(t)$ is a time varying function that monotonically increases from 0 to 1.
As $f(t)\rightarrow1$, the approximate distance function gradually transforms from an initial function, $\phi_i(\bm{x})$, to a final function, $\phi_f(\bm{x})$.
For $f(t)$, we use a modified hyperbolic tangent function:
\begin{equation}
    f(t)=\frac{e^{pt}-1}{e^{pt}+1}
    \label{eq:modified_tanh_function}
\end{equation}
where $p$ is a positive constant ($p>0$) that dictates the speed at which the function $f(t)$ transforms from 0 to 1, and thus the speed at which the robot will transform from one shape to another. 

In summary, Eqs.\ \ref{eq:transfinite_morph} and \ref{eq:modified_tanh_function} provide a way to smoothly transition a distance function from an initial shape, $S_{i}$, to a final shape, $S_{f}$, in a controlled manner.
The next section will discuss the simulation and experimental platforms that the distance functions and space time transfinite interpolation were applied to with the objective of forming a desired shape and morphing between shapes.  

\section{Material and Methods}
\label{sec:Material and Methods}
This section describes the control algorithm, simulation environment, and experimental platform used to verify the approach. 
\begin{figure}
    \centering
    \includegraphics[width=0.48\textwidth]{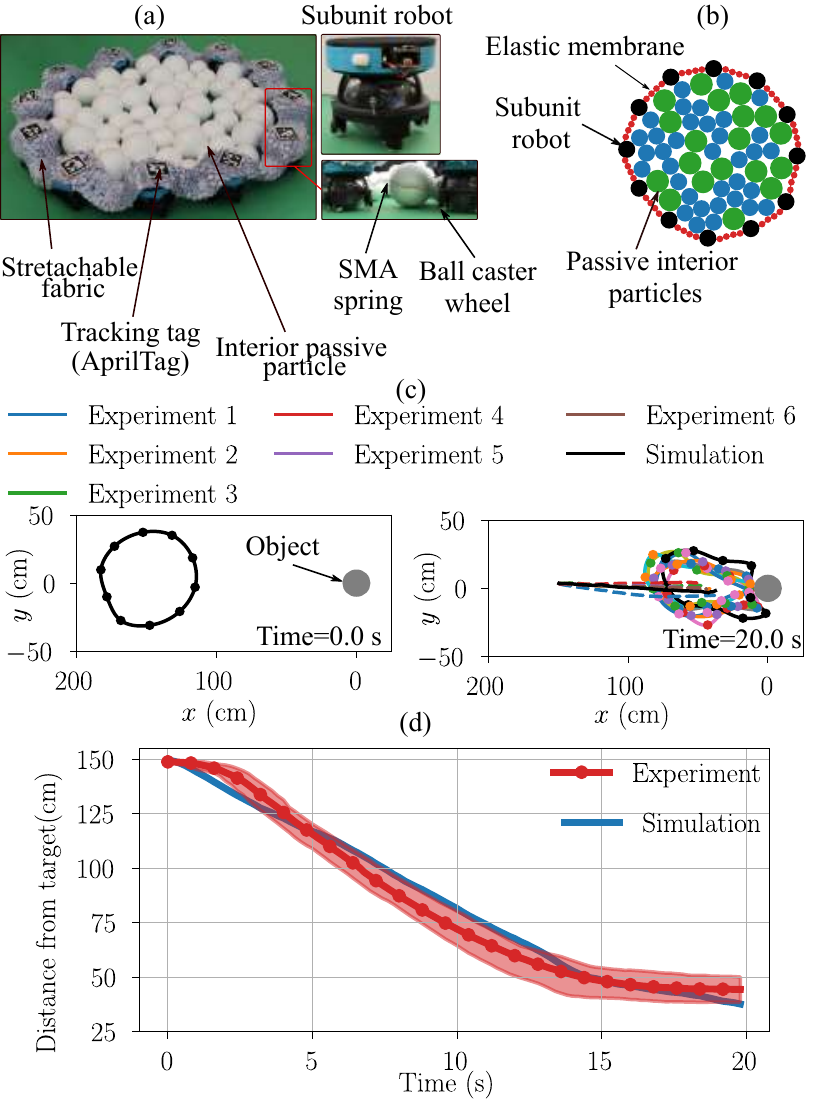}
    \caption{(a): The experimental two-dimensional planar boundary-constrained soft robot. (b): The simulated system. (c): Snapshots of the simulated and six experimental systems pursuing a target.(d): Distance from the target for the simulated and the average of the six experimental systems.} 
    \label{fig:experimental_system}
\end{figure}
\subsection{Experimental Platform}
\label{sec:Experimental Platform}
The experimental robot (see Fig.~\ref{fig:experimental_system}(a)) is a two-dimensional planar representation of the general system shown in Fig.\ \ref{fig:General_Concept_v2}.  
The robot consists of ten omnidirectional sub-robots that each use a commercial Sphero Bolt to provide locomotion.
Each sub-robot is flexibly connected to each other through shape memory alloy springs (Kellog's Research Labs, \SI{0.5}{mm} wire size, \SI{9.5}{mm} mandrel size, tight pitch, \SI{35}{\degreeCelsius} transition temperature, spring coefficient $k=$\SI{6.4}{\newton \per \meter} off and $k=$\SI{13.8}{\newton \per \meter} on) and extensible fabric (Gilbins B07S395K3J). 
The shape memory alloy springs help the robot execute a jamming function, a feature not utilized in this work, which only focuses on shape formation. 
Interior particles consist of Styrofoam passive spheres (Crafjie \SI{76.2}{\milli\meter} and \SI{101.6}{\milli\meter} foam balls), roughly the same size of the active subunits.

Experiments were performed on a paper-covered table (Savage SAV461253) with an overhead camera (Logitec Brio 960-001105) that visually tracked each sub-robot's pose in real time using AprilTags \cite{wang2016apriltag}.
A central computer running ROS computed all desired control inputs, which were broadcast to the subunits via Bluetooth.
Although the system contains rigid bodies, it exhibits the continuous nature and configurable properties of a soft robot. 

\subsection{Simulation}\label{sec:Simulation}
The robot model consists of three components: active boundary sub-robots, passive particles in the interior, and an elastic membrane. 
The membrane is approximated through a series of spring-mass systems, both in 2D and 3D.
See Fig.~\ref{fig:experimental_system}(b).
The passive interior and robots are modeled as rigid bodies.
For the passive interior we used a mixture of particles with radii of $r_{1}$=\SI{3.25}{\centi\meter}, and $r_{2}=\sqrt{2}r_{1}$.
The inclusion of particles with different radii was done to prevent crystallization, which is a phenomenon that emerges in granular media with identical particles~\cite{OHern2001}.
The properties used were:
sub-robot (radius \SI{3}{\centi\meter}, mass \SI{200}{\gram}, number 30), interior particle (radius \SI{3.25}{\centi\meter}, mass = \SI{30}{\gram}, number 180), friction coefficient 0.2, spring stiffness \SI{50}{\newton\per\meter}.

The system's equation motion is expressed as:
\begin{equation} \label{eq:CD_global}
    \mathbf{M}\Ddot{\bm{x}} +\mathbf{K}\bm{x} = \bm{u}+\mathbf{F}_c
\end{equation}
where $n$ is the number of rigid bodies, $\mathbf{M} \in\mathbb{R}^{3n\times 3n}$ is the  system's generalized mass matrix, and $\mathbf{K} \in\mathbb{R}^{3n\times 3n}$ is a matrix representing the interconnected springs between the boundary particles and active robots.
$\bm{x} \in\mathbb{R}^{3n\times 1}$ is a column vector containing the degrees of freedom of the system ($x,y,\theta$ for each rigid body), $\bm{u} \in\mathbb{R}^{3n\times 1}$ contains the control inputs to all degree of freedoms (zero for passive degrees of freedom).
The column vector, $\mathbf{F}_c \in\mathbb{R}^{3n\times 1}$, contains the contact forces that emerge as the constituting bodies interact with each other. 
The model was simulated in the open source physics engine Project Chrono \cite{Chrono2016}.

To verify the simulation platform, we simulated the experimental robot moving towards a desired target and compared it to six experimental tests performing the same task (see Fig.\ \ref{fig:experimental_system}(c)).
Fig.\ \ref{fig:experimental_system}(d) compares the distance from the robot's center of mass to the target of the experimental and simulated systems with respect to time. 
Both Fig.\ \ref{fig:experimental_system}(c) and (d) show good alignment.

\subsection{Control}
The control law for the $i^\mathrm{th}$ active boundary sub-robot is given as:
\begin{equation}
\bm{u}_{i}=-{\alpha}{\nabla}\phi(\bm{x})\big\vert_{\bm{x}=\bm{q}_{i}} 
\label{eq:controller}
\end{equation}
where $\bm{u}_{i}$ is the system's control input, ${\nabla}\phi(\bm{x})$ is the gradient of the distance function corresponding to the desired shape,  $\bm{q}_{i}=[x_{i},y_{i}]$ is the position vector of the $i^\mathrm{th}$ sub-robot, and $\alpha$ is sub-robot's thrust. 
Eq.~\ref{eq:controller} is commonly used in other potential field control laws~\cite{Hsieh2006potfieldsshape,chaimowicz2005controlling,choset2005principles}; 
The difference here is that the function used to calculate the gradient, $\phi(\bm{x})$, is an approximate distance function rather than a potential field.
Note that only one distance function, $\phi$, is generated for shape formation objectives and that two distance functions are generated for shape morphing objectives. The field, $\phi_{i}$, corresponding to the initial shape, and the field, $\phi_{f}$, corresponds to the desired shape.
Additionally, we replace $\phi({\bm{x}}$) with $\phi({\bm{x},t})$ as shown in Eq.~\ref{eq:transfinite_morph}.

\section{Results}\label{sec:Shape-forming}
In this section, we present experimental and simulation results that demonstrate that the approximate distance functions can be used to control the shape of a soft robot.  
The demonstrations include forming a single shape and transitioning from one shape to another for the two-dimensional simulation and experimental platform.  
We then extend this to demonstrate how the approach would work for a three-dimensional system in simulation only.  
Note for both the simulations and experiments the robots have no knowledge of the other robots, only their own position.


\begin{figure}
    \centering
    \includegraphics[width=0.48\textwidth]{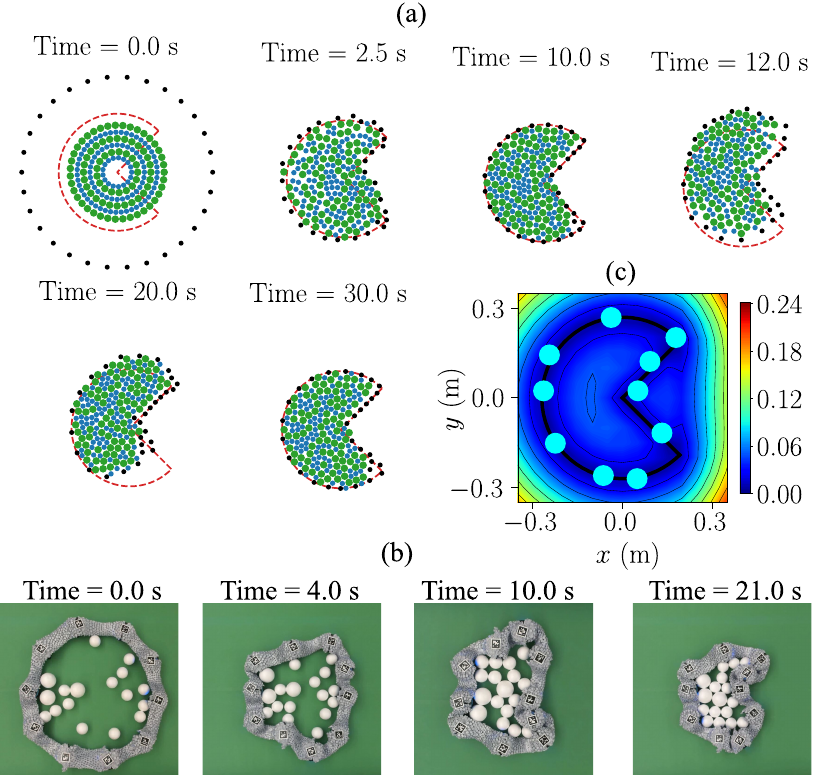}
    \caption{(a): The simulated system forming a Pac-Man shape (see Movie S1). (b): Snap shots of the experimental system transitioning shapes (see Movie S1). (c): Distance function utilized for the experimental system with the final positions of the experimental sub-robots overlaid.}
    \label{fig:pacman}
\end{figure}

Fig.\ \ref{fig:pacman}(a) shows a simulated system forming and maintaining a Pac-Man like shape.
Starting at \SI{10}{\second} and ending at \SI{15}{\second}, a series of external disturbances are applied to the system, but the system is shown to recover.
Fig.\ \ref{fig:pacman}(c) shows snapshots of the experimental system forming a similar shape. 
The distance function was created using R-equivalence of two line segments and an arc.

\subsection{Morphing}\label{subsec:Morphing}
\begin{figure}
    \centering
    \includegraphics[width=0.48\textwidth]{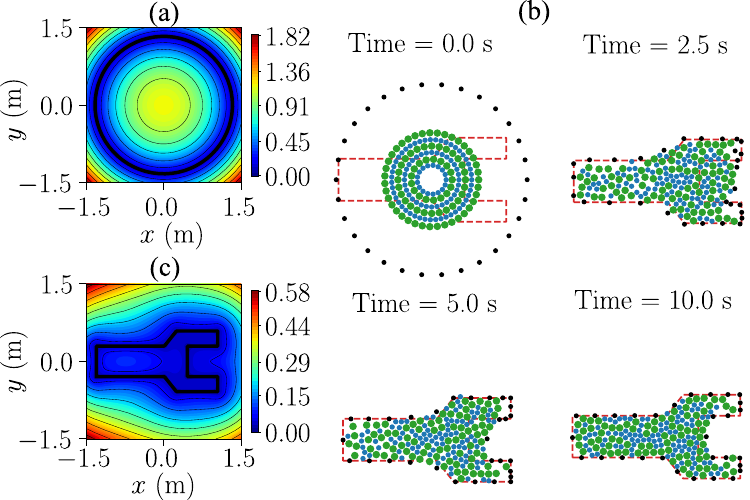}
    \caption{(a): Approximate distance function of the initial shape, $S_{i}$. (b): Approximate distance function of the desired shape, $S_{f}$. (c): Snapshots of the simulated system transitioning shapes using space time transfinite interpolation (see Movie S1).}
    \label{fig:wrench}
\end{figure}
Fig.\ \ref{fig:wrench} shows an example of the system morphing from a circular configuration, $S_{i}$, to a wrench-shaped configuration, $S_{f}$, using the approximate distance function shown in Fig.\ \ref{fig:wrench}(c).

\subsection{3-D Shape Forming}\label{subsec:3-D morphing}
\begin{figure}
    \centering
    \includegraphics[width=0.48\textwidth]{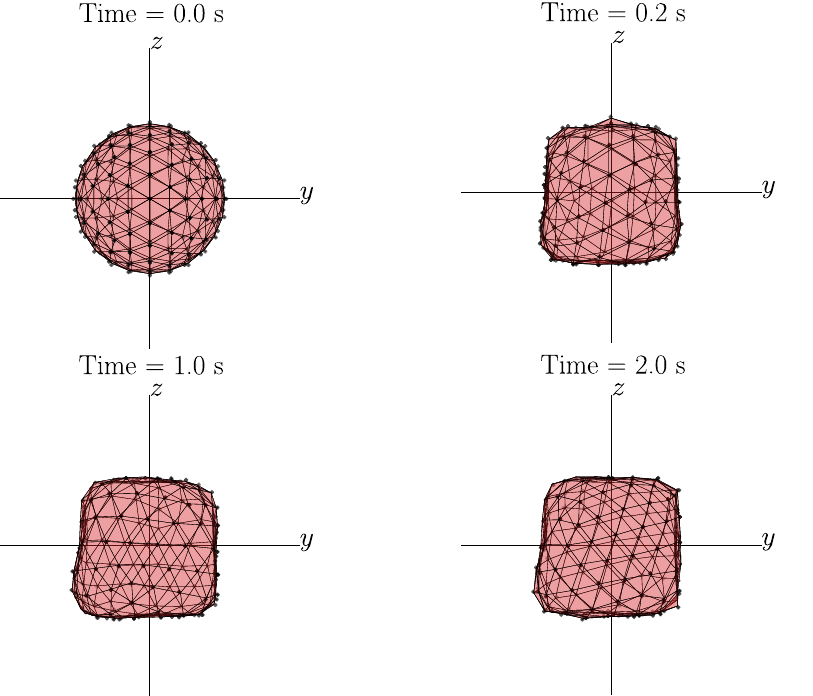}
    \caption{A 3D JAMoEBA system forming a cube. Here we had 162 boundary robots with 800 interior particles}
    \label{fig:cube}
\end{figure}
This section demonstrates how distance functions might be used to control shape formation for a three-dimensional soft robot assuming the sub-robots have omnidirectional locomotion capabilities. 
To simulate this system we used an alternative platform, PyBullet~\cite{coumans2020}, for its ability to easily interface rigid bodies with soft structures.
This allowed us to simulate a membrane that encapsulates the passive interior particles.
The system is modeled with an icosphere mesh where each of the 162 nodes represents an active boundary sub-robot.
The control algorithm enacts a shape formation by forming a cube (see Fig.~\ref{fig:cube}). 
The approximate distance functions were constructed by performing R-disjunction, Eq.~\ref{eq:R_alpha}, with a series of approximate distance functions for a plane, Eq.~\ref{eq:plane}.

\section{Discussion, Conclusion, and Future Work}
\label{sec:discussion}
In this paper, we present a control methodology incorporating distance functions, R-functions, and transfinite interpolation to control soft robotic systems. 
We demonstrated these techniques on soft robots based on boundary-constrained granular swarms in both experiments and simulation.
An advantage of this technique is that the robot's elastic outer membrane and granular interior implicitly enforce distance constraints.
This is in contrast to most traditional swarm robot systems in which each robot needs knowledge of its neighbors poses to form a desired shape.

The control methods based on R-functions may be compared with more well-known potential field-based methods. 
The main advantage of the method presented here is the easy evaluation of the surface $\phi$ for arbitrarily complex desired shapes of the soft robot in terms of analytical compositions of elementary functions. 
The resulting $\phi$ has guaranteed differentiability properties leading to further stability in the controller. An important step in the proposed method is the creation of $\phi$ for complex shapes, preferably in an automated fashion. 
This requires the division of the composite shape into more elementary sub-shapes (e.g., arcs, lines, surfaces and planes). 
A promising alternative might be to exploit nurbs to define complex shapes for which R-functions are already available~\cite{Yang2019SplineRA}.
One could subsequently use the equivalence operation to fully automate the process of the creation of $\phi$.

One potential issue, as it pertains to the present experimental platform, is the emergence of localized jamming when a shape transformation is attempted too quickly.
This can be mitigated by slowing the speed of the transfinite interpolation or by introducing small amplitude random oscillations in the robot forces.
Future work will need to be conducted to mitigate localizing jamming of the system.
Additionally, we will also apply these techniques for grasping and transporting objects.

\section*{Acknowledgments}
We thank Vahid Alizadehyazdi, David Canones, and Koki Tanaka for assistance with the experimental system.
The authors have no competing interests. 
All data needed to support the conclusions are found in the manuscript and supplementary materials. 
All code for simulation and experiments can be found on (https://github.com/dmulroy/R-functions.git)


\begin{thebibliography}{10}
\providecommand{\url}[1]{#1}
\csname url@samestyle\endcsname
\providecommand{\newblock}{\relax}
\providecommand{\bibinfo}[2]{#2}
\providecommand{\BIBentrySTDinterwordspacing}{\spaceskip=0pt\relax}
\providecommand{\BIBentryALTinterwordstretchfactor}{4}
\providecommand{\BIBentryALTinterwordspacing}{\spaceskip=\fontdimen2\font plus
\BIBentryALTinterwordstretchfactor\fontdimen3\font minus
  \fontdimen4\font\relax}
\providecommand{\BIBforeignlanguage}[2]{{%
\expandafter\ifx\csname l@#1\endcsname\relax
\typeout{** WARNING: IEEEtran.bst: No hyphenation pattern has been}%
\typeout{** loaded for the language `#1'. Using the pattern for}%
\typeout{** the default language instead.}%
\else
\language=\csname l@#1\endcsname
\fi
#2}}
\providecommand{\BIBdecl}{\relax}
\BIBdecl

\bibitem{Sanchez2015}
M.~Sanchez, O.~Fryazinov, V.~Adzhiev, P.~Comninos, and A.~A. Pasko,
  ``Space-time transfinite interpolation of volumetric material properties,''
  \emph{IEEE Transactions on Visualization and Computer Graphics}, vol.~21, pp.
  278--288, 2015.

\bibitem{pottmann2005industrial}
H.~Pottmann, S.~Leopoldseder, M.~Hofer, T.~Steiner, and W.~Wang, ``Industrial
  geometry: recent advances and applications in cad,'' \emph{Computer-Aided
  Design}, vol.~37, no.~7, pp. 751--766, 2005.

\bibitem{persson2004simple}
P.-O. Persson and G.~Strang, ``A simple mesh generator in matlab,'' \emph{SIAM
  review}, vol.~46, no.~2, pp. 329--345, 2004.

\bibitem{osher2003signed}
S.~Osher and R.~Fedkiw, ``Signed distance functions,'' in \emph{Level set
  methods and dynamic implicit surfaces}.\hskip 1em plus 0.5em minus
  0.4em\relax Springer, 2003, pp. 17--22.

\bibitem{Fossel2015}
\emph{{2D-SDF-SLAM: A signed distance function based SLAM frontend for laser
  scanners}}, 2015.

\bibitem{Billard_DS_planning}
L.~Huber, A.~Billard, and J.-J. Slotine, ``Avoidance of convex and concave
  obstacles with convergence ensured through contraction,'' \emph{IEEE Robotics
  and Automation Letters}, vol.~4, no.~2, pp. 1462--1469, 2019.

\bibitem{jeremias2013shadertoy}
P.~Jeremias and I.~Quilez, ``Shadertoy: live coding for reactive shaders,'' in
  \emph{ACM SIGGRAPH 2013 Computer Animation Festival}.\hskip 1em plus 0.5em
  minus 0.4em\relax Association for Computing Machinery, 2013, pp. 1--1.

\bibitem{Shapiro:1991}
V.~Shapiro, ``Theory of {R-functions} and applications: {A} primer,'' Cornell
  Programmable Automation, Sibley School of Mechanical Engineering, Ithaca, NY
  14853, USA, Tech. Rep. CPA88-3, 1991.

\bibitem{tanaka2020cable}
K.~Tanaka, M.~A. Karimi, B.-P. Busque, D.~Mulroy, Q.~Zhou, R.~Batra,
  A.~Srivastava, H.~M. Jaeger, and M.~Spenko, ``Cable-driven jamming of a
  boundary constrained soft robot,'' in \emph{2020 3rd IEEE International
  Conference on Soft Robotics (RoboSoft)}, 2020, pp. 852--857.

\bibitem{karimi2020boundary}
M.~A. Karimi, V.~Alizadehyazdi, B.-P. Busque, H.~M. Jaeger, and M.~Spenko, ``A
  boundary-constrained swarm robot with granular jamming,'' in \emph{2020 3rd
  IEEE International Conference on Soft Robotics (RoboSoft)}, 2020, pp.
  291--296.

\bibitem{Karimi2021}
M.~A. Karimi, V.~Alizadehyazdi, H.~M. Jaeger, and M.~Spenko, ``A
  self-reconfigurable variable-stiffness soft robot based on
  boundary-constrained modular units,'' \emph{IEEE Trans Robotics}, vol. 2021,
  pp. 1--12, 2021.

\bibitem{agrawal2020scale}
M.~Agrawal and S.~C. Glotzer, ``Scale-free, programmable design of morphable
  chain loops of kilobots and colloidal motors,'' \emph{Proc National Academy
  of Sciences}, vol. 117, no.~16, pp. 8700--8710, 2020.

\bibitem{chaimowicz2005controlling}
L.~Chaimowicz, N.~Michael, and V.~Kumar, ``{Controlling swarms of robots using
  interpolated implicit functions},'' \emph{IEEE Int Conf Robotics and
  Automation}, vol. 2005, no. April, pp. 2487--2492, 2005.

\bibitem{Ge2004}
S.~S. Ge, C.~H. Fua, and W.~M. Liew, ``{Swarm formations using the general
  formation potential function},'' \emph{2004 IEEE Conference on Robotics,
  Automation and Mechatronics}, no.~2, pp. 655--660, 2004.

\bibitem{payne1992}
B.~Payne and A.~Toga, ``Distance field manipulation of surface models,''
  \emph{IEEE Computer Graphics and Applications}, vol.~12, no.~1, pp. 65--71,
  1992.

\bibitem{Borgefors1996}
G.~Borgefors, ``{On digital distance transforms in three dimensions},''
  \emph{Computer Vision and Image Understanding}, vol.~64, no.~3, pp. 368--376,
  1996.

\bibitem{Shapiro:2007:SAG}
V.~Shapiro, ``Semi-analytic geometry with {R}-functions,'' \emph{Acta
  Numerica}, vol.~16, pp. 239--303, 2007.

\bibitem{Rvachev:1995:RBV}
V.~L. Rvachev and T.~I. Sheiko, ``R-functions in boundary value problems in
  mechanics,'' \emph{amr}, vol.~48, no.~4, pp. 151--188, 1995.

\bibitem{Keshmiri2018}
S.~Keshmiri, A.~R. Kim, D.~Shukla, A.~Blevins, and M.~Ewing, ``{Flight Test
  Validation of Collision and Obstacle Avoidance in Fixed-Wing UASs with High
  Speeds Using Morphing Potential Field},'' \emph{2018 Int Conf Unmanned
  Aircraft Systems, ICUAS 2018}, pp. 589--598, 2018.

\bibitem{Shapiro1999}
V.~Shapiro and I.~Tsukanov, ``{Implicit functions with guaranteed differential
  properties},'' \emph{Proceedings of the Symposium on Solid Modeling and
  Applications}, pp. 258--269, 1999.

\bibitem{Millan2015:CBM}
D.~Millán, N.~Sukumar, and M.~Arroyo, ``Cell-based maximum-entropy
  approximants,'' \emph{Computer Methods in Applied Mechanics and Engineering},
  vol. 284, pp. 712--731, 2015.

\bibitem{Biswas:2004:ADF}
A.~Biswas and V.~Shapiro, ``Approximate distance fields with non-vanishing
  gradients,'' \emph{Graphical Models}, vol.~66, no.~3, pp. 133--159, 2004.

\bibitem{wang2016apriltag}
J.~Wang and E.~Olson, ``Apriltag 2: Efficient and robust fiducial detection,''
  in \emph{IROS}.\hskip 1em plus 0.5em minus 0.4em\relax IEEE, 2016, pp.
  4193--4198.

\bibitem{OHern2001}
C.~S. O'Hern, S.~A. Langer, A.~J. Liu, and S.~R. Nagel, ``{Force distributions
  near jamming and glass transitions},'' \emph{Physical Review Letters},
  vol.~86, no.~1, pp. 111--114, 2001.

\bibitem{Chrono2016}
A.~Tasora, R.~Serban, H.~Mazhar, A.~Pazouki, D.~Melanz, J.~Fleischmann,
  M.~Taylor, H.~Sugiyama, and D.~Negrut, ``Chrono: An open source multi-physics
  dynamics engine,'' in \emph{High Performance Computing in Science and
  Engineering}, T.~Kozubek, Ed.\hskip 1em plus 0.5em minus 0.4em\relax
  Springer, 2016, pp. 19--49.

\bibitem{Hsieh2006potfieldsshape}
M.~Y.~A. Hsieh and V.~Kumar, ``{Pattern generation with multiple robots},''
  \emph{Proceedings - IEEE International Conference on Robotics and
  Automation}, vol. 2006, no. May, pp. 2442--2447, 2006.

\bibitem{choset2005principles}
H.~M. Choset, K.~M. Lynch, S.~Hutchinson, G.~Kantor, W.~Burgard, L.~Kavraki,
  S.~Thrun, and R.~C. Arkin, \emph{Principles of robot motion: theory,
  algorithms, and implementation}.\hskip 1em plus 0.5em minus 0.4em\relax MIT
  press, 2005.

\bibitem{coumans2020}
E.~Coumans and Y.~Bai, ``Pybullet, a python module for physics simulation for
  games, robotics and machine learning,'' \url{http://pybullet.org},
  2016--2020.

\bibitem{Yang2019SplineRA}
T.~Yang, A.~Qarariyah, and J.~Deng, ``Spline r-function and applications in
  fem,'' \emph{Numerical Mathematics: Theory, Methods and Applications},
  vol.~13, no.~1, pp. 150--175, 2019.

\end{thebibliography}

\end{document}